\documentclass{article}
\usepackage{ijcai23}

\usepackage{times}
\usepackage{soul}
\usepackage{url}
\usepackage[hidelinks]{hyperref}
\usepackage[utf8]{inputenc}
\usepackage[small]{caption}
\usepackage{graphicx}
\usepackage{amsmath}
\usepackage{amsthm}
\usepackage{booktabs}
\usepackage{algorithm}
\usepackage{algorithmic}
\usepackage[switch]{lineno}
\usepackage{graphicx}
\usepackage{latexsym}
\usepackage{times}
\usepackage{soul}
\usepackage{url}
\usepackage[hidelinks]{hyperref}
\usepackage[utf8]{inputenc}
\usepackage{graphicx}
\usepackage{amsmath}
\usepackage{amsthm}
\usepackage{booktabs}
\usepackage{algorithm}
\usepackage{algorithmic}
\usepackage{times}
\usepackage{soul}
\usepackage{url}
\usepackage[hidelinks]{hyperref}
\usepackage[utf8]{inputenc}
\usepackage{graphicx}
\usepackage{amsmath}
\usepackage{amssymb}
\usepackage{booktabs}
\usepackage{algorithm}
\usepackage{algorithmic}
\usepackage[switch]{lineno}
\usepackage{color}
\usepackage{array}
\usepackage{booktabs}
\usepackage{multirow}
\usepackage{bm}
\usepackage{mathrsfs}
\usepackage{enumitem}
\usepackage{amsthm}

\title{Modeling Selective Feature Attention for\\ Representation-based Siamese Text Matching}

\author{
Jianxiang Zang
\and
Hui Liu\textsuperscript{\thanks{Corresponding author}}
\affiliations
School of Statistics and Information, \\Shanghai University of International Business and Economics\\
\emails
\{21349110, liuh\}@suibe.edu.cn,
}

\begin{document}

\maketitle

\begin{abstract}
Representation-based Siamese networks have risen to popularity in lightweight text matching due to their low deployment and inference costs. While word-level attention mechanisms have been implemented within Siamese networks to improve performance, we propose \textbf{F}eature \textbf{A}ttention (\textbf{FA}), a novel downstream block designed to enrich the modeling of dependencies among embedding features. Employing "squeeze-and-excitation" techniques, the FA block dynamically adjusts the emphasis on individual features, enabling the network to concentrate more on features that significantly contribute to the final classification. Building upon FA, we introduce a dynamic "selection" mechanism called \textbf{S}elective \textbf{F}eature \textbf{A}ttention (\textbf{SFA}), which leverages a stacked BiGRU Inception structure. The SFA block facilitates multi-scale semantic extraction by traversing different stacked BiGRU layers, encouraging the network to selectively concentrate on semantic information and embedding features across varying levels of abstraction. Both the FA and SFA blocks offer a seamless integration capability with various Siamese networks, showcasing a plug-and-play characteristic. Experimental evaluations conducted across diverse text matching baselines and benchmarks underscore the indispensability of modeling feature attention and the superiority of the "selection" mechanism.~\footnote{Codes available:https://github.com/hggzjx/SFA}
\end{abstract}


\section{Introduction}

The goal of the text matching task is to assess the semantic relevance between pairs of sentences and to determine their relationship. More specifically, it involves creating a classifier \(\xi\) that calculates the conditional probability \(P(\text{label}|\bm{s}^{a},\bm{s}^{b})\), thereby predicting the relationship between the sentence pair \(\bm{s}^{a}\) and \(\bm{s}^{b}\). Here, \(\text{label} \in \Omega\) represents different levels of sentence pair relationships, which can be \{relevant, irrelevant\} or \{entailed, neutral, contradicted\}. Representation-based Siamese networks~\cite{wang2017bilateral,chen2017enhanced,yang2019simple,zang2023extract} use dual encoders to compute text embeddings offline and aggregate them downstream for prediction. They offer the benefits of having low parameter sizes and reduced inference latency, making them extensively applicable in industrial contexts, including search engines and recommendation systems~\cite{huang2013learning,khattab2020colbert}. To enhance the post-interaction of text pairs, researchers have introduced various downstream attentions in Siamese networks~\cite{chen2017enhanced,yang2019simple,cao2020deformer,liu2021pay}. 
Notably, these attention strategies solely capture \textbf{\emph{word-level}} dependencies, neglecting the modeling of intricate relationships among \textbf{\emph{embedding features}}. Each feature in text embeddings captures certain semantic or syntactic properties of the vocabulary, though these properties are usually not directly interpretable. For example, particular dimensions in the embedding vector might be related to parts of speech, contextual information (the relationship of a word to surrounding words), or other linguistic attributes.

\begin{figure*}[t]
\centering
\includegraphics[width=0.8\textwidth]{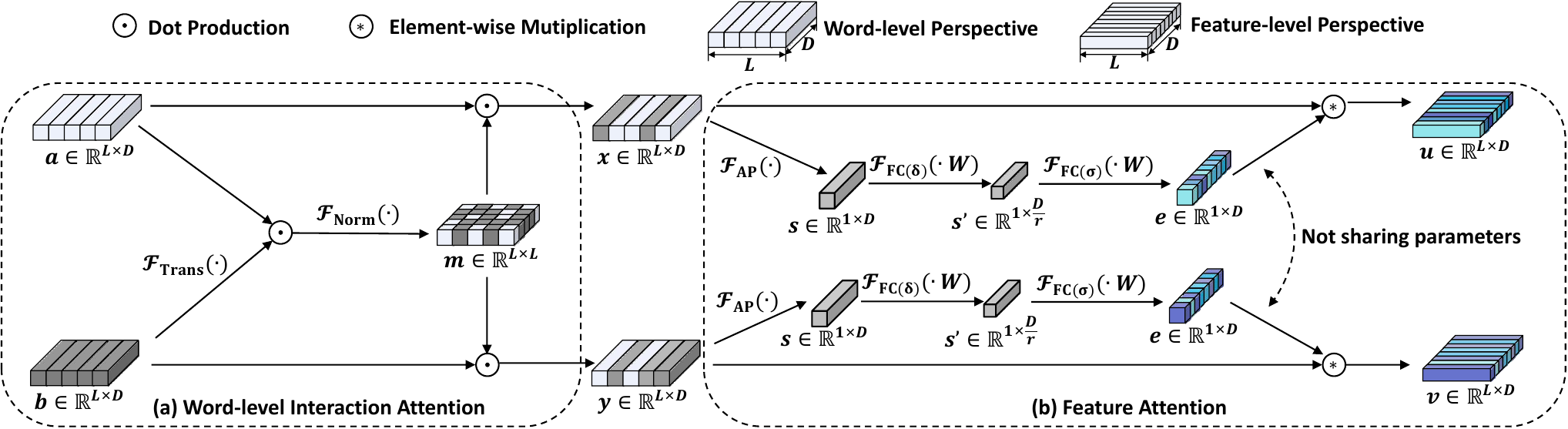}
\caption{Our more robust downstream attention, composed of (a) Word-level Interaction Attention and (b) Feature Attention.
}\label{fig.fa}
\end{figure*}

The Word-level Interaction Attention shown in Figure \ref{fig.fa}(a) is the most commonly used downstream attention in Siamese matching networks~\cite{chen2017enhanced,tay2018compare,yang2019simple}, and is a mapping \((\bm{a},\bm{b})\rightarrow(\bm{x},\bm{y})\), where \(\bm{a},\bm{b},\bm{x}, \bm{y} \in \mathbb{R}^{L \times D}\). Here, \(\bm{a},\bm{b}\) represent the text embeddings of the text pair \(\bm{s}^{a},\bm{s}^{b}\). \(\bm{x},\bm{y}\) represent the embeddings of the text pair containing rich word level interaction information. \(L\) denotes the length of the text sequence, and \(D\) represents the dimension of the embedding features. To enhance the sensitivity of the Siamese network to the embedding features then construct a more robust downstream attention, as illustrated in Figure \ref{fig.fa}(b), we advocate further building a single-branch symmetric Feature Attention (FA) based on the Word-level Interaction Attention. The FA block is a mapping that does not change the tensor size: \(\bm{x}\rightarrow \bm{u}\) or \(\bm{y}\rightarrow \bm{v}\), where \(\bm{u}, \bm{v} \in \mathbb{R}^{L \times D}\). It is noteworthy that, despite the FA blocks in both branches sharing the \emph{same} form, we advocate \emph{against} sharing their parameters.

The FA block incorporates a "squeeze-and-excitation" approach, which concentrates on the most influential embedding features, enhancing their significance in the final classification.
Moreover, inspired by neuroscience, where the size of receptive fields in visual cortical neurons is modulated by external stimuli, we integrate a dynamic "selection" mechanism into Feature Attention based on the stacked BiGRU Inception structure. This results in the creation of Selective Feature Attention (SFA). The SFA block stimulates the network to dynamically adapt its focus on semantic information and embedding features across various levels of abstraction. Simultaneously, the "selection" within SFA effectively addresses the challenge of consistent gradient flow arising from the diverse scale semantic extraction in multi-branch Inception, achieving more efficient gradient flow management. 

The FA and SFA blocks preserve tensor shape in their mappings, capable of seamless integration with virtually any Siamese network, offering a plug-and-play characteristic. 
In the experimental evaluation, we combine the FA and SFA blocks with six of the most commonly used baseline Siamese networks from 2020 to 2023, assessing their performance across various text matching benchmarks. Extensive experiments demonstrate that the integration of SFA with all networks significantly improves inference accuracy across all text matching benchmarks.
Our primary contributions are highlighted in the following: (1)Based on our survey, we are the first to model dependencies at the embedding feature level for text matching. (2)We present the Feature Attention block and enhance it with a "selection" mechanism based on the stacked BiGRU Inception structure, resulting in the Selective Feature Attention. Extensive experiments confirm the superior performance of the "selection" in SFA block. (3)FA and SFA blocks are offering a plug-and-play characteristic, allowing them to be integrated with almost any Siamese network.

\begin{figure*}[h]
\centering
\includegraphics[width=0.95\textwidth]{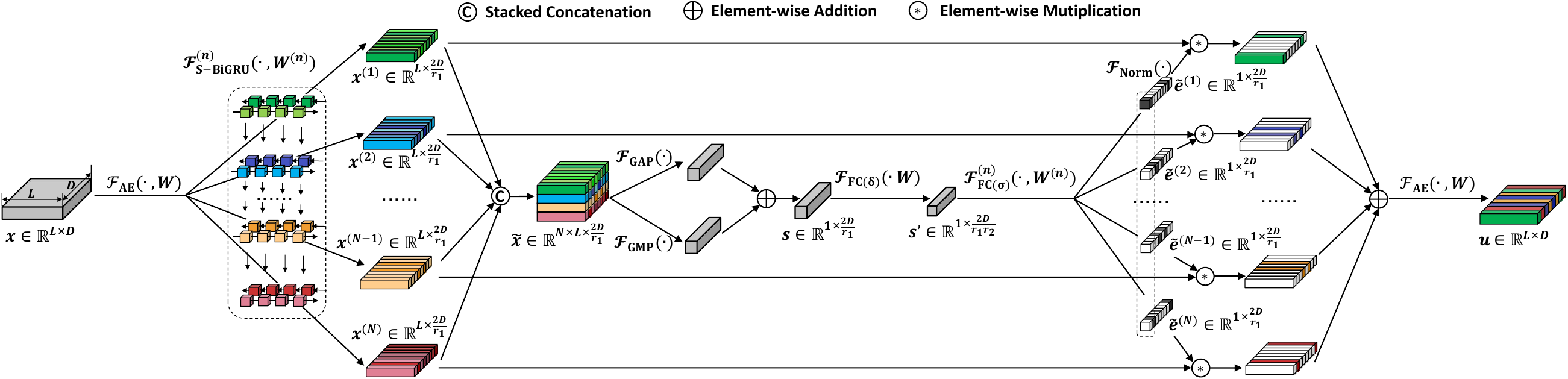}
\caption{Selective Feature Attention
}\label{fig.sfa}
\end{figure*}

\section{Feature Attention}

\subsection{Squeeze-and-Excitation Network}


Channel-level attention mechanisms have demonstrated exceptional performance in image classification~\cite{hu2018squeeze,hu2018gather,woo2018cbam,bello2019attention} and segmentation~\cite{hou2020strip,huang2019ccnet,fu2019dual}. The Squeeze-and-Excitation Network (SE-Net)~\cite{hu2018squeeze} pioneered channel attention by effectively constructing interdependencies among channels through the compression of each feature map. CBAM~\cite{woo2018cbam} further refined this concept by introducing spatial information encoding via convolutions with large-size kernels. Subsequent studies such as GENet~\cite{hu2018gather}, GALA~\cite{linsley2019learning}, TA~\cite{misra2021rotate} expanded on this idea by adopting various spatial attention mechanisms or designing advanced attention blocks. While related work~\cite{zang2023improving} models feature dependencies through higher-dimensional semantic spaces, we devised a squeeze-and-excitation style Feature Attention to model dependencies among semantic features.

\subsection{FA Block}



As illustrated in Figure~\ref{fig.fa}(b), FA block constitutes a 2D to 2D single-branch mapping that computes $\bm{x}\rightarrow \bm{u}$ or $\bm{y}\rightarrow \bm{v}$. For the sake of simplicity, our discussion will focus solely on the computational mapping of the $\bm{x}$ branch. 
 
For the input \(\bm{x} \in \mathbb{R}^{L \times D}\), to capture feature-level dependencies, we first execute a "squeeze" step using average pooling (\(\mathcal{F}_{\text{AP}}(\cdot)\)) to compress global information into a feature descriptor \(\bm{s}\). Formally, \(\bm{s} \in \mathbb{R}^{1 \times D}\) is generated by averaging \(\bm{x}\) along the spatial dimension \(L\), where the \(d^{\text{th}}\) element of \(\bm{s}\) is computed as formulated in Equation~\ref{eq.ap}. Throughout this paper, $\mathcal{F}(\cdot)$ represents the mapping that does not involve trainable weights, while $\mathcal{F}(\cdot,\bm{W})$ represents the mapping that involves trainable weights \(\bm{W}\). Symbols with the subscript \(\cdot_l\) denote spatial (word-level) descriptors, while those with the subscript \(\cdot_d\) represent feature descriptors.

\setlength{\abovedisplayskip}{1pt}
\begin{equation}
    \bm{s}_d = \frac{1}{L}\sum_{l=1}^L\bm{x}_l\quad,\quad d \in [1,...,D]\label{eq.ap}
\end{equation}

In the subsequent "excitation" step, aimed at enhancing the model's sensitivity to features, we filter out the embedding features that contribute more significantly to the final classification. For the aggregated information \(\bm{s}\) from the previous steps, the "excitation" step is tasked with constructing a nonlinear, non-mutually exclusive gating mechanism. To ensure the excitation of multiple features, we have designed two fully connected layers for nonlinear mapping (\(\mathcal{F}_{\text{FC}(\delta)}(\cdot,\bm{W})\) and \(\mathcal{F}_{\text{FC}(\sigma)}(\cdot,\bm{W})\)), namely a dimension-reducing layer with a Tanh function followed by a dimension-increasing layer with a Sigmoid function. Here, \(\delta\), \(\sigma\) represent the Tanh and Sigmoid function respectively, and \(\bm{s}' \in \mathbb{R}^{1 \times \frac{D}{r}}\) is a transitional vector. The decay factor \(r\) introduces a bottleneck in the network to control parameter redundancy. 


\setlength{\abovedisplayskip}{1pt}
\begin{equation}
    \bm{e} = \sigma(\delta(\bm{s}\bm{W}_{\text{FC}_1})\bm{W}_{\text{FC}_2})
\end{equation}


The vector \(\bm{e} \in \mathbb{R}^{1 \times D}\) signifies the features that have been activated. Ultimately, this vector is merged with \(\bm{x}\) through element-wise multiplication, enabling \(\bm{x}\) to further develop into an embedding feature representation \(\bm{u} = [\bm{u}_1, \bm{u}_2,..., \bm{u}_D] \in \mathbb{R}^{L \times D}\) that is more finely attuned to the final classification.

\setlength{\abovedisplayskip}{1pt}
\begin{equation}
    \bm{u}_d = \bm{e}_d*\bm{x}_d\quad,\quad d \in [1,...,D]
\end{equation}

\section{Selective Feature Attention}

\subsection{Inception Structure}




In the visual cortex, neurons' ability to gather multi-scale spatial information within the same processing stage stems from the varying receptive field sizes in the same region~\cite{hubel1962receptive}. The Inception structure~\cite{szegedy2017inception} leverages this characteristic, achieving superior performance in computer vision by directly concatenating features extracted from multiple scales. However, this linear aggregation may be insufficient to model the neurons' robust adaptability. Furthermore, the uniform treatment of semantic extraction at different scales gradient flow management during training, affecting training stability.

Related research has shown that stimuli also influence neuronal responses~\cite{nelson1978orientation}. The size of these receptive fields is not fixed but correlates with the stimulus contrast: lower contrast corresponds to a larger effective receptive field~\cite{sceniak1999contrast}. Selective Kernel Networks (SK-Net)~\cite{li2019selective} were the first to model this phenomenon in the field of computer vision, achieving significant success in image classification and semantic segmentation.
Motivated by this theory, we introduce the Selective Feature Attention, which encourages the network to dynamically adapt its focus on semantic information and embedding features across different levels of abstraction.

\subsection{SFA Block}


The SFA block comprises three phases: "split-and-fusion", "squeeze-and-excitation", and "selection", as illustrated in Figure~\ref{fig.sfa}. Initially, considering the potential complexity introduced by a multi-branch Inception structure, we advocate the incorporation of bottleneck structure at both ends of the SFA block for feature dimension scaling. Specifically, we employ a one-dimensional convolution kernel of size 1 as an auto encoder \(\mathcal{F}_{\text{AE}}(\cdot,\bm{W})\) to map the size of the input \(\bm{x}\) to \(\mathbb{R}^{L \times \frac{D}{r_1}}\), where \(r_1\) acts as a dimension reduction factor controlling the feature dimensionality of embeddings.


In the "split-and-fusion" phase, for \(\bm{x} \in \mathbb{R}^{L \times \frac{D}{r_1}}\), we introduce an \(N\)-layer stacked BiGRU ($\mathcal{F}_{\text{S-BiGRU}}(\cdot,\bm{W}^{(n)})$) to capture the semantic representation at each layer, effectively "splitting" the original embedding into vectors \(\{\bm{x}^{(n)}\}_{n=1}^N\), with \(\bm{x}^{(n)} \in \mathbb{R}^{L \times \frac{2D}{r_1}}\), as formulated in Equation~\ref{eq.split}. Here, \(h_l^{(n)}\) represents the hidden state at position \(l\) in the \(n^{\text{th}}\) layer of either the forward or backward GRU and $<;>$ denotes concatenation along the feature dimension.

\begin{equation}
\begin{aligned}
    \bm{x}_l^{(n)} = <&\overrightarrow{\text{GRU}}^{(n)}(\overrightarrow{h}_{l-1}^{(n)}, \bm{x}^{(n-1)}_l,\bm{W}^{(n)}_{\overrightarrow{\text{GRU}}});\\&\overleftarrow{\text{GRU}}^{(n)}(\overleftarrow{h}_{l+1}^{(n)}, \bm{x}^{(n-1)}_l,\bm{W}^{(n)}_{\overleftarrow{\text{GRU}}})>\\&,\small{n \in [1,...,N],l \in [1,...,L]}\label{eq.split}  
\end{aligned}
\end{equation}



The shallower layers of BiGRU excel at capturing short-range dependencies between words, such as understanding the combination of words in compound words or phrases. On the other hand, deeper layers of BiGRU are capable of processing and capturing long-range word dependencies. This includes discerning a sentence's theme, which may hinge on words at the beginning and end of the sentence or require consideration of the entire sentence's content for accurate interpretation.  The core idea behind "fusion" is to use a gating mechanism to enable information carrying different levels of semantic abstraction from multiple branches to flow towards the neurons of the next layer. To comprehensively and holistically preserve the semantic information of each branch, we employ stacked concatenation to amalgamate the results of all branches, as formulated in Equation~\ref{eq.fuse}. Here $[;]$ denotes the stacked concatenation.

\setlength{\abovedisplayskip}{1pt}
\begin{equation}
    \widetilde{\bm{x}}^{(n)}_l=[\bm{x}_l^{(1)};\bm{x}_l^{(2)};...;\bm{x}_l^{(n)}], \small{\quad n \in [1,...,N],l \in [1,...,L]}~\label{eq.fuse}
\end{equation}


In the subsequent "squeeze-and-excitation" phase, for \(\widetilde{\bm{x}} \in \mathbb{R}^{N \times L \times \frac{2D}{r_1}}\), we recommend the combined use of global average pooling ($\mathcal{F}_{\text{GAP}}(\cdot)$) and global max pooling ($\mathcal{F}_{\text{GMP}}(\cdot)$) to  compress information simultaneously at both the BiGRU layer and word levels. As formulated in Equation~\ref{eq.ap2}, we obtain the sum \(\bm{s}\) from the global average pooling and max pooling results, and then apply fully connected layers for activation.

\setlength{\abovedisplayskip}{1pt}
\begin{equation}
    \bm{s}_d = \frac{1}{N\times L}\sum_{n=1}^N\sum_{l=1}^L\widetilde{\bm{x}}^{(n)}_l+\underset{n=1}{\overset{N}{\text{max}}}\underset{l=1}{\overset{L}{\text{max}}} (\widetilde{\bm{x}}^{(n)}_l),\small{\quad d \in [1,...,\frac{2D}{r_1}]}\label{eq.ap2}
\end{equation}


It is noteworthy that the essence of SFA block is to capture and excite features of text embeddings at different levels of abstraction, while adaptively adjust their relative importance. To accomplish this, we employ a single dimension-reducing fully connected layer (\(\mathcal{F}_{\text{FC}(\delta)}(\cdot,\bm{W})\)) alongside a series of dimension-increasing fully connected layers, with the count matching the number of branches (\(\mathcal{F}_{\text{FC}(\sigma)}(\cdot,\bm{W}^{(n)})\)) for excitation. This process results in the excited vectors \(\{\bm{e}^{(n)}\}_{n=1}^N, \bm{e}^{(n)} \in \mathbb{R}^{1 \times \frac{2D}{r_1}}\), as formulated in Equation~\ref{eq.e3}. Similarly, \(\delta\),\(\sigma\) represent the Tanh, Sigmoid function, respectively. The decay factor \(r_2\) creates a bottleneck in the network to avoid parameter redundancy.

\setlength{\abovedisplayskip}{1pt}
\begin{equation}
    \bm{e}^{(n)} = \sigma(\delta(\bm{s}\bm{W}_{\text{FC}_1})\bm{W}^{(n)}_{\text{FC}_2}))\quad,\quad n\in[1,...,N]\label{eq.e3}
\end{equation}


In the most critical "selection" phase, we apply vector-level softmax normalization (\(\mathcal{F}_{\text{Norm}}(\cdot)\)) to the excitation vectors \(\{\bm{e}^{(n)}\}_{n=1}^N\). The normalized results \(\{\widetilde{\bm{e}}^{(n)}\}_{n=1}^N\) serve as adaptive weights for the different branches, and are used for element-wise multiplication with each branch's representation \(\{\bm{x}^{(n)}\}_{n=1}^N\). The summation of these products yields the weighted sum representation of each branch \(\bm{u}=[\bm{u}_1,\bm{u}_2...,\bm{u}_{\frac{2D}{r_1}}] \in \mathbb{R}^{L \times \frac{2D}{r_1}}\).

\begin{small}
\begin{equation}
\begin{aligned}
        \bm{u}_d =\sum_{n=1}^N\widetilde{\bm{e}}^{(n)}_d*\bm{x}^{(n)}_d = \frac{\sum_{n=1}^N\text{exp}(\bm{e}^{(n)}_d)*\bm{x}^{(n)}_d}{\sum_{n=1}^N\text{exp}(\bm{e}^{(n)}_d)},d\in [1,..., \frac{2D}{r_1}]
\end{aligned}
\end{equation}     
\end{small}

Finally, to align the input and output dimensions of the SFA block, we also employ the same auto encoder (\(\mathcal{F}_{\text{AE}}(\cdot)\)) to transform the shape of \(\bm{u}\) to \(\mathbb{R}^{L \times D}\). To ensure the expressive power of the embedding feature, the feature dimension reduction caused by the bottleneck layer in the entire text must satisfy~\cite{lai2016generate}:


\setlength{\abovedisplayskip}{1pt}
\begin{small}
\begin{equation}
    \frac{D}{r},\frac{2D}{r_1},\frac{2D}{r_1r_2}>8.33\log L
\end{equation}
\end{small}

\begin{table*}[t]
\centering
\small
\setlength{\tabcolsep}{5 pt} 
\renewcommand{\arraystretch}{0.85} 
\begin{tabular}{l|cc|llllllll|cc}
\hline
\textbf{Network} & $\bm{r}$ & $\bm{N}$ & \textbf{QQP}   & \textbf{MRPC}  & \textbf{BoolQ} & \textbf{SNLI}  & \textbf{MNLI(m/mm)}    & \textbf{QNLI}  & \textbf{Scitail} & \textbf{Avg.}  & \textbf{P\textsubscript{avg.}(M)} & \textbf{IL\textsubscript{avg.}(ms)} \\\hline
BiMPM            & -                                  & -                                  & 85.54          & 70.38          & 69.75          & 87.22          & 72.34/72.02          & 79.24          & 79.45            & 76.99          & 1.83                                               & 0.403                                                \\
+FA            & 2                                     & 1                                     & 85.79          & 70.61\dag          & 70.18\dag          & 87.19          & 72.74\dag/72.41\dag          & 80.19\dag          & 80.13\dag            & 77.41\dag          & 2.01                                               & 0.481                                                \\
+SFA           & (3.75, 4)                              & 2                                     & \textbf{86.13\dag} & \textbf{71.33\dag} & \textbf{70.86\dag} & \textbf{88.14\dag} & \textbf{73.19\dag/72.97\dag} & \textbf{80.94\dag} & \textbf{80.98\dag}   & \textbf{78.17\dag} & 2.06                                               & 0.708                                                \\\hline
ESIM             & -                                  & -                                  & 87.92          & 73.48          & 71.71          & 88.04          & 74.27/74.19          & 80.84          & 79.23            & 78.71          & 4.46                                               & 0.672                                                \\
+FA            & 1                                     & 1                                     & 88.38\dag          & 73.59          & 72.32\dag          & 88.59\dag          & 74.28/74.11         & 81.16\dag          & 80.17\dag            & 79.08\dag          & 4.82                                               & 0.718                                                \\
+SFA           & (3,5)                               & 3                                     & \textbf{90.32\dag} & \textbf{75.88\dag} & \textbf{73.94\dag} & \textbf{90.02\dag} & \textbf{76.31\dag/76.19\dag} & \textbf{82.92\dag} & \textbf{81.62\dag}   & \textbf{80.90\dag} & 4.88                                               & 1.248                                                \\\hline
CAFE             & -                                  & -                                  & 88.01          & 73.18          & 71.23          & 87.68          & 75.02/74.45          & 81.65          & 80.54            & 78.97          & 4.75                                               & 0.672                                                \\
+FA            & 1                                     & 1                                     & 89.04\dag          & 73.25          & 71.34          & 87.98\dag          & 75.34\dag/74.47          & 81.71          & 81.75            & 79.36          & 5.11                                               & 0.708                                                \\
+SFA           & (3,5)                               & 3                                     & \textbf{90.27\dag} & \textbf{74.54\dag} & \textbf{73.21\dag} & \textbf{89.72\dag} & \textbf{77.03\dag/76.33\dag} & \textbf{82.84\dag} & \textbf{82.82\dag}   & \textbf{80.85\dag} & 5.17                                               & 1.386                                                \\\hline
RE2              & -                                  & -                                  & 88.78          & 73.17          & 72.11          & 88.29          & 75.98/75.52          & 80.36          & 82.45            & 79.58          & 4.85                                               & 0.742                                                \\
+FA            & 1                                     & 1                                     & 90.14\dag          & 73.51\dag          & 72.97\dag          & 88.87\dag          & 76.11/75.71\dag          & 80.53\dag          & 82.38            & 80.03\dag          & 5.21                                               & 0.805                                                \\
+SFA           & (3,5)                               & 3                                     & \textbf{90.97\dag} & \textbf{75.29\dag} & \textbf{74.28\dag} & \textbf{90.52\dag} & \textbf{77.61\dag/77.41\dag} & \textbf{81.97\dag} & \textbf{84.52\dag}   & \textbf{81.57\dag} & 5.27                                               & 1.478                                                \\\hline
DIIN             & -                                  & -                                  & 88.26          & 73.03          & 71.45          & 88.08          & 76.56/76.49          & 81.67          & 82.34            & 79.74          & 4.42                                               & 0.653                                                \\
+FA            & 1                                     & 1                                     & 88.71\dag          & 73.42\dag          & 71.78\dag          & 88.56\dag          & 76.43/76.37         & 82.05\dag          & 82.87\dag            & 80.02\dag          & 4.78                                               & 0.689                                                \\
+SFA           & (3,5)                               & 3                                     & \textbf{90.34\dag} & \textbf{75.04\dag} & \textbf{73.73\dag} & \textbf{89.33\dag} & \textbf{78.03\dag/77.82\dag} & \textbf{82.68\dag} & \textbf{84.51\dag}   & \textbf{81.44\dag} & 4.84                                               & 1.289                                                \\\hline
DRCN             & -                                  & -                                  & 88.81          & 72.45          & 71.67          & 89.84          & 78.07/77.85          & 81.03          & 82.98            & 80.34          & 6.68                                               & 1.436                                                \\
+FA            & 1                                     & 1                                     & 90.18          & 72.56          & 71.98\dag          & 90.25\dag          & 78.38\dag/78.17\dag          & 81.21          & 83.07            & 80.73\dag          & 7.40                                               & 1.608                                                \\
+SFA           & (3,5)                               & 3                                     & \textbf{90.53\dag} & \textbf{74.34\dag} & \textbf{73.57\dag} & \textbf{90.96\dag} & \textbf{79.25\dag/78.95\dag} & \textbf{82.38\dag} & \textbf{84.97\dag}   & \textbf{81.87\dag} & 7.51                                               & 2.803\\ \hline                                               
\end{tabular}
\caption{The evaluation accuracy (\%) of introducing FA and SFA on 6 lightweight text matching baselines across 7 text matching benchmarks. The hyperparameters $\bm{r}$ and $\bm{N}$ represent the dimension reduction factors and the number of branches in the Inception network, respectively. P\textsubscript{avg.}(M) denotes the model's average parameters (million), and IL\textsubscript{avg.}(ms) indicates the sentence-level inference latency. The bolded parts represent the best values in each group, and \dag  signifies a significant improvement over the baseline (t-test, $p<0.05$).}~\label{tab.main}
\end{table*}

\subsection{Efficient Gradient Management}\label{sec.grad}



The traditional Inception structure aggregates multi-scale information linearly.
However, this uniform updating of weights across different Inception layers impedes the differentiated flow of gradients, thereby impacting training stability. The distinct mapping branches $\mathcal{F}^{(n)}_{\text{S-BiGRU}}(\cdot ,\bm{W}^{(n)}): \bm{x} \rightarrow \bm{x}^{(n)}$ represent semantic extraction at various scales from the text embeddings, acknowledging that semantic extraction at different scales contributes differently to the final classification. In this section, we analyze the backpropagation within the SFA block during training, focusing on how the "selection" mechanism impacts training stability in the context of the gradient flow $\frac{\partial \bm{u}}{\partial \bm{x}}$.

\noindent {\textbf{w/o selection}} Firstly, an SFA block without "selection" implies that there is only a single feature weight \(\bm{e}\), thus we have $\bm{u} = \bm{e} * \bm{\widetilde{x}}$. Consequently, the gradient flow chain is jointly determined by the gradient propagations of both $\bm{e}$ and $\bm{\widetilde{x}}$, specifically $\frac{\partial \bm{u}}{\partial \bm{\widetilde{x}}} = \bm{e}$ and $\frac{\partial \bm{u}}{\partial \bm{e}} = \bm{\widetilde{x}}$. The gradient flow in the direction of $\bm{e}$ is formulated in Equation~\ref{eq.e}. Since the process involves a straightforward addition, it results in $\frac{\partial \bm{s}}{\partial \mathcal{F}_{\text{GMP}}(\widetilde{\bm{x}})} = 1$, $\frac{\partial \bm{s}}{\partial \mathcal{F}_{\text{GAP}}(\widetilde{\bm{x}})} = 1$.

\begin{small}
\begin{equation}
\begin{aligned}
       &\frac{\partial \bm{u}}{\partial \widetilde{\bm{x}}} =  \frac{\partial \bm{u}}{\partial \bm{e}} * \frac{\partial \bm{e}}{\partial \bm{s}'} * \frac{\partial \bm{s}'}{\partial \bm{s}}* (\frac{\partial \mathcal{F}_{\text{GMP}}(\widetilde{\bm{x}})}{\partial \widetilde{\bm{x}}}+\frac{\partial \mathcal{F}_{\text{GAP}}(\widetilde{\bm{x}})}{\partial \widetilde{\bm{x}}})\label{eq.e} 
\end{aligned}
\end{equation}
\end{small}


Combining the backward propagation of gradients from $\bm{x}\rightarrow \bm{\widetilde{x}}$, the computation of the overall gradient flow is formulated in Equation~\ref{eq.grad1}. In this context, the transformation from $\bm{x}^{(n)} \rightarrow \widetilde{\bm{x}}$ is a stacked concatenation, hence it follows that $\frac{\partial \widetilde{\bm{x}}}{\partial \bm{x}^{(n)}} = 1$. 

\setlength{\abovedisplayskip}{1pt}
\begin{small}
\begin{equation}
\begin{aligned}
        \frac{\partial \bm{u}}{\partial \bm{x}}=2*\frac{\partial \bm{u}}{\partial \widetilde{\bm{x}}}*\sum_{n=1}^N\frac{\partial \widetilde{\bm{x}}}{\partial \bm{x}^{(n)}}*\frac{\partial \bm{x}^{(n)}}{\partial \bm{x}}=2*\frac{\partial \bm{u}}{\partial \widetilde{\bm{x}}}*\sum_{n=1}^N\frac{\partial \bm{x}^{(n)}}{\partial \bm{x}}\label{eq.grad1}
\end{aligned}
\end{equation}
\end{small}


It can be observed that each semantic extraction mapping \(\mathcal{F}^{(n)}_{\text{S-BiGRU}}: \bm{x} \rightarrow \bm{x}^{(n)}\) represents a gradient flow \(\frac{\partial \bm{x}^{(n)}}{\partial \bm{x}}\), which impacts the overall gradient flow \(\frac{\partial \bm{u}}{\partial \bm{x}}\) in a uniform proportion of \(2*\frac{\partial \bm{u}}{\partial \widetilde{\bm{x}}}\). This implies that the feature weights of each BiGRU layer are updated to the same degree. However, this uniformity does not align with the necessity for differentiated gradient flow management across various semantic scales, resulting in unstable training.


\noindent {\textbf{w/ selection}} The "selection" mechanism of the SFA block introduces an adaptive weight for each branch, leading to the entire gradient flow chain being jointly determined by the gradient propagations of both $\bm{x}^{(n)}$ and $\widetilde{\bm{e}}^{(n)}$. This is expressed as $\frac{\partial  \bm{u}}{\partial \bm{x}^{(n)}}=\bm{e}^{(n)}$ and $\frac{\partial \bm{u}}{\partial \widetilde{\bm{e}}^{(n)}}=\bm{x}^{(n)}$. The gradient flow in the direction of $\widetilde{\bm{e}}^{(n)}$ is formulated in Equation~\ref{eq.en}. Since the process involves a straightforward addition, it results in $\frac{\partial \bm{u}}{\partial \widetilde{\bm{e}}^{(n)}} = 1$, as formulated in Equation~\ref{eq.en}.

\setlength{\abovedisplayskip}{1pt}
\begin{small}  
\begin{equation}
\begin{aligned}
    \frac{\partial \bm{u}}{\partial \widetilde{\bm{x}}} =& \sum_{n=1}^N ( \frac{\partial \bm{u}}{\partial \widetilde{\bm{e}}^{(n)}} *\frac{\partial \widetilde{\bm{e}}^{(n)}}{\partial \bm{e}^{(n)}} * \frac{\partial \widetilde{\bm{e}}^{(n)}}{\partial \bm{s}'} ) 
    \\&* \frac{\partial \bm{s}'}{\partial \bm{s}} * (\frac{\partial \mathcal{F}_{\text{GMP}}(\widetilde{\bm{x}})}{\partial \widetilde{\bm{x}}}+\frac{\partial \mathcal{F}_{\text{GAP}}(\widetilde{\bm{x}})}{\partial \widetilde{\bm{x}}})\label{eq.en}
\end{aligned}
\end{equation}
\end{small}


After backpropagating the gradients in Equation~\ref{eq.en}, a complete gradient chain is formulated in Equation~\ref{eq.p2}. It is evident that, upon the introduction of a selection mechanism, the coefficient preceding each $\frac{\partial \bm{x}^{(n)}}{\partial \bm{x}}$ becomes a linear function of $\widetilde{\bm{e}}^{(n)}$. This equation reflects how $\frac{\partial \bm{x}^{(n)}}{\partial \bm{x}}$ directly and, through the adaptive weights $\widetilde{\bm{e}}^{(n)}$, indirectly influences the total gradient flow. In contrast to the scenario without the "selection", where $\frac{\partial \bm{x}^{(n)}}{\partial \bm{x}}$ affects the total gradient flow uniformly, the adaptive weights $\widetilde{\bm{e}}^{(n)}$ manage the gradient flow across different branches in an adaptive and differentiated manner. This approach leads to a more robust training process, as reflected in Section~\ref{sec.ablation}.

\setlength{\abovedisplayskip}{1pt}
\begin{small}
\begin{equation}
\begin{aligned}
        \frac{\partial \bm{u}}{\partial \bm{x}}&=\sum_{n=1}^N(\frac{\partial  \bm{u}}{\partial \bm{x}^{(n)}}*\frac{\partial \bm{x}^{(n)}}{\partial \bm{x}})+\frac{\partial  \bm{u}}{\partial \widetilde{\bm{x}}}*\sum_{n=1}^N(\frac{\partial \widetilde{\bm{x}}}{\partial \bm{x}^{(n)}}*\frac{\partial \bm{x}^{(n)}}{\partial \bm{x}})\\&=\sum_{n=1}^N(\widetilde{\bm{e}}^{(n)}+\frac{\partial  \bm{u}}{\partial \widetilde{\bm{x}}})*\frac{\partial \bm{x}^{(n)}}{\partial \bm{x}}\label{eq.p2}
\end{aligned}
\end{equation}
\end{small}

\section{Experimental Results \& Analysis}

\subsection{Main Results}

We selected the six most commonly used lightweight baselines in text matching tasks from 2020 to 2023. All these baselines reported parameter sizes in the articles with values less than ten million. The selected baselines are: BiMPM~\cite{wang2017bilateral}, ESIM~\cite{chen2017enhanced}, CAFE~\cite{tay2018compare}, RE2~\cite{yang2019simple}, DIIN~\cite{gong2018natural}, and DRCN~\cite{kim2019semantic}. In the experiments, we introduce FA block and SFA block to these baseline networks and evaluated their performance on following benchmarks: QQP~\cite{iyer2017first}, MRPC~\cite{dolan2005automatically}, BoolQ~\cite{clark2019boolq}, SNLI\cite{bowman2015large}, MNLI~\cite{williams2018broad}(matched\&mismatched), QNLI~\cite{wang2018glue}, and Scitail~\cite{khot2018scitail}.

Table~\ref{tab.main} reports the evaluation accuracies of six lightweight text matching baselines, as well as their performances following the integration of FA and SFA blocks. Particularly, since DRCN is composed of multiple identical modules in series, we incorporated two FA and SFA blocks in each branch to align with the number of autoencoders present in DRCN. To minimize the impact of increased parameters on performance, we carefully controlled the values of $r$ and $N$ to maintain a consistent increment in parameters caused by the FA and SFA blocks. By regulating these hyperparameters, the additional parameters introduced by FA and SFA amounted to approximately 5\%-10\% of the network's original parameters. The sentence-level inference latency caused by the SFA block is higher than that of the FA block, due to the increased model complexity resulting from the stacked BiGRU. The incorporation of the FA block lead to overall improvements across all baselines, while the introduction of the SFA block significantly boost inference accuracy. The SFA block consistently demonstrate the best performance across all baselines compared to the FA block. It provides the most substantial average performance boost for the ESIM model (from 78.71\% to 80.90\%). With the integration of the SFA block, DRCN achieved the highest accuracy (81.87\%) among all baselines, albeit at the cost of having the highest network parameters and inference latency.

\begin{table}[t]
\renewcommand{\arraystretch}{0.80} 
\centering
\small
\begin{tabular}{lcccc}
\hline
\textbf{Network} & \textbf{QQP}   & \textbf{SNLI}  & \textbf{P\textsubscript{avg.}(M)} & \textbf{IL\textsubscript{avg.}(ms)}\\ \hline
Residual Stacked* & -           & 86.0             & 29                                                 & -                    \\ 
LSTM-Max*         & -           & 84.5           & 40                                                 & -                    \\ 
LM-Transformer*   & -           & 89.9           & 85                                                 & -                    \\ 
AlBERT           & 89.31          & 87.30          & 12                                                 & 7.27                    \\ 
BERT-base        & 90.06         & 90.16          & 109                                                & 7.74                    \\ 
BERT-large       & 90.45          & 90.82 & 335                                                & 27.26                    \\ 
RoBERTa-base          & \textbf{90.73} & \textbf{91.13}          & 125                                                & 12.05                    \\ \hline
BiPMP-SFA        & 86.13          & 88.14          & 2.06                                                & 0.71                    \\ 
ESIM-SFA         & 90.32          & 90.02          & 4.88                                                & 1.25                    \\ 
CAFE-SFA         & 90.27         & 89.72          & 5.17                                                & 1.39                    \\ 
RE2-SFA          & \textbf{90.97}          & 90.52          & 5.27                                                & 1.48                    \\ 
DIIN-SFA         & 90.34          & 89.33          & 4.84                                                & 1.29                    \\ 
DCRN-SFA         & 90.53          & \textbf{90.96} & 7.51                                                & 2.80                    \\ \hline
\end{tabular}
\caption{Evaluation results of the network with the SFA block integrated, compared with other large-scale networks on QQP and SNLI. * signifies that the result directly adopts the accuracy as reported in~\protect\cite{kim2019semantic}.The bolded parts represent the two best results for each evaluation benchmark.}
\label{tab.llms}
\end{table}

Table~\ref{tab.llms} reports a comprehensive comparison of six SFA-enhanced lightweight baselines against several high-parameter networks, (Residual Stacked~\cite{nie2017shortcut}, LSTM-Max~\cite{conneau2017supervised}, LM-Transformer~\cite{vaswani2017attention}, BERT-base/large~\cite{devlin2019bert}, AlBERT~\cite{lan2019albert}, and RoBERTa~\cite{liu2019roberta}). The comparison is based on evaluation accuracies, parameter sizes, and inference latencies across the QQP and SNLI benchmarks. In the QQP benchmark, it is observed that after integrating SFA, networks such as ESIM, CAFE, RE2, DIIN, and DRCN not only retain a substantial advantage in parameter volume and inference latency but also surpass the performance of pre-trained models like AlBERT (89.31\%) and BERT (90.06\%). Particularly noteworthy is RE2-SFA, which, with only 1.5\% and 4.2\% of the parameters and 5.4\%, 12.3\% of the inference latency, surpasses the accuracy of BERT-large (90.45\%) and RoBERTa-base (90.73\%) at 90.97\%. In the SNLI benchmark, our networks with SFA outperform Residual Stacked, LSTM-Max, LM-Transformer, and AlBERT in terms of accuracy. Notably, DRCN, with just 6.9\% and 2.2\% of the parameter size and 36.2\%, 10.3\% of the inference latency, surpasses both BERT-base (90.16\%) and BERT-large (90.82\%). Despite having only 6.0\% of the parameters and 23.2\% of the inference latency, DRCN closely approaches the accuracy of RoBERTa-base (91.13\%).




\subsection{Ablation Study}\label{sec.ablation}

To investigate the key factors contributing to the superiority of the SFA block over the FA block, we conducted detailed ablation studies on the various components of the SFA block. As indicated in Table~\ref{tab.main}, ESIM and RE2 demonstrated the most significant average performance improvement following the integration of the SFA block. Thus, for a clearer representation of the results, we chose these two networks, along with QQP and SNLI, as the baselines and benchmarks for our ablation experiments. 

\begin{figure}[t]
  \centering
  \includegraphics[width=1\linewidth]{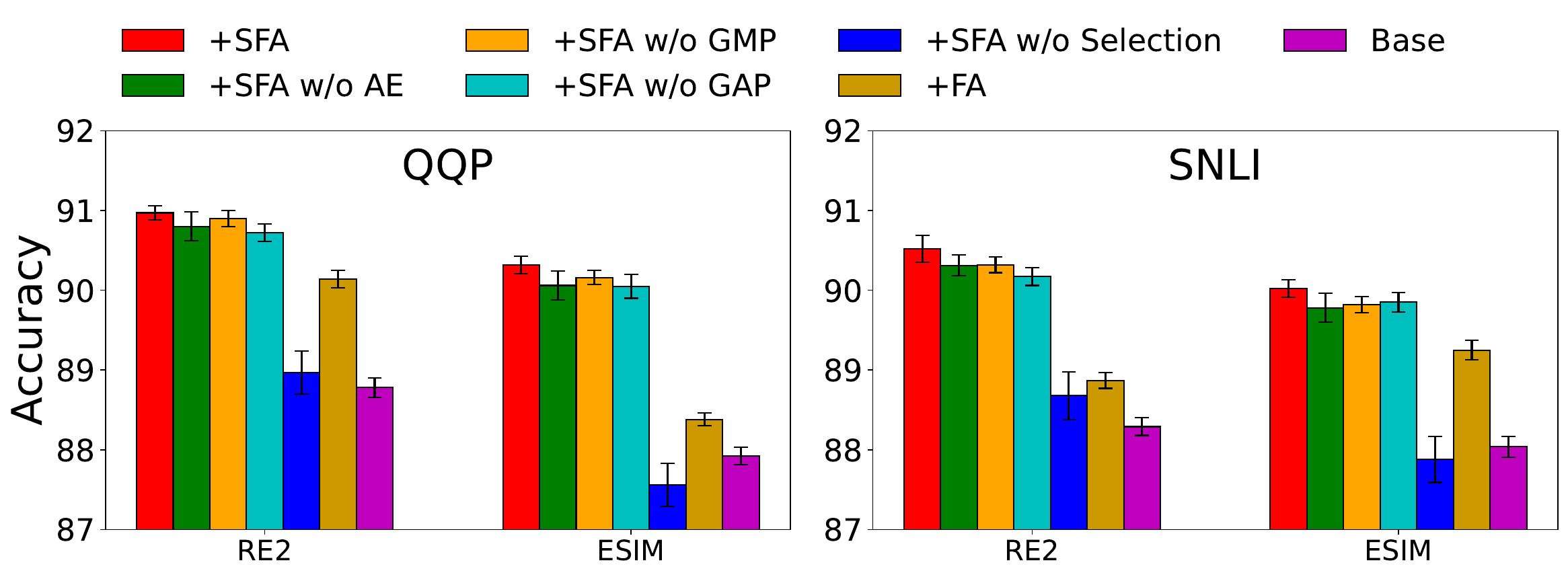}
  \includegraphics[width=1\linewidth]{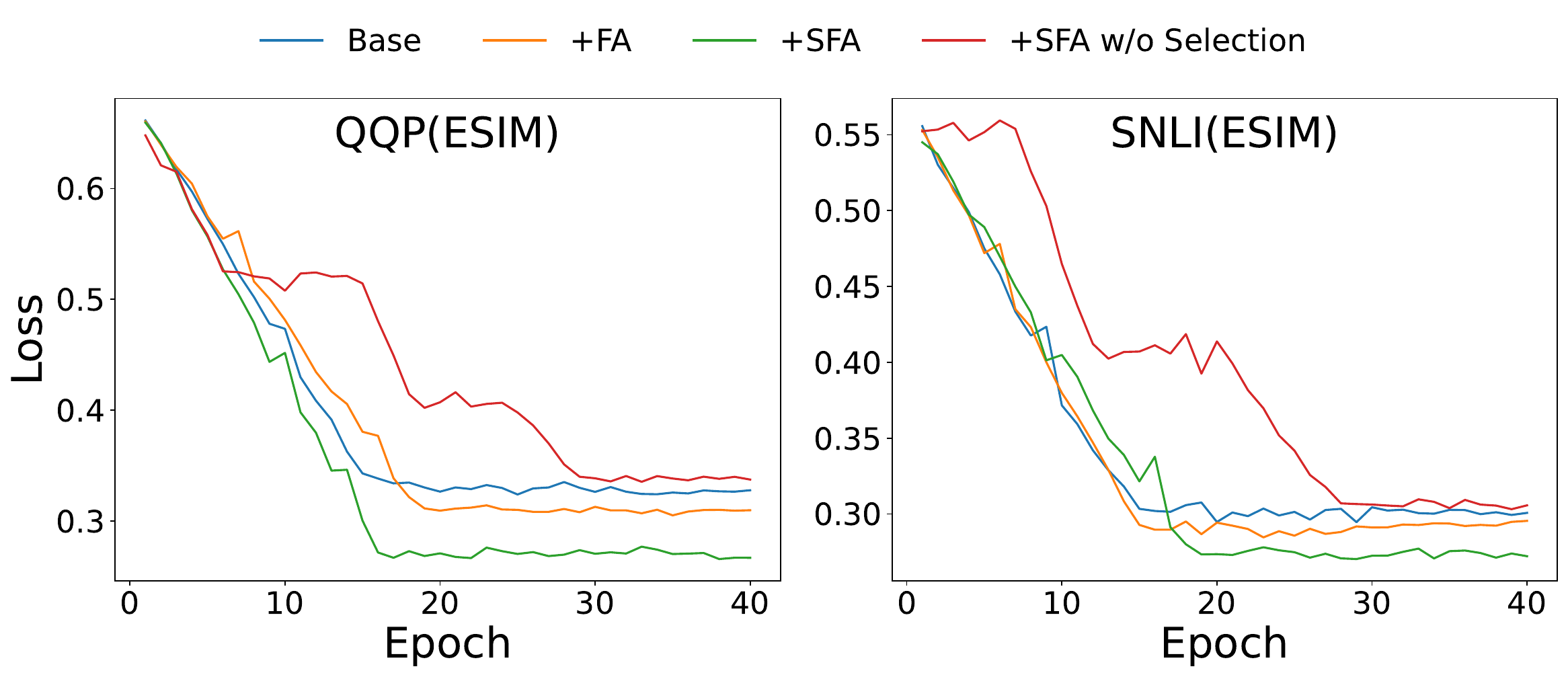}
  \caption{Ablation study on the components of SFA block on QQP and SNLI datasets, using RE2 and ESIM as baselines.}
 \label{fig.ablation}
\end{figure}

Figure~\ref{fig.ablation} illustrates the changes in prediction accuracy when components such as auto encoder, global max pooling, global average pooling, and the "selection" within the SFA block are individually removed. It is evident that the exclusion of each of these components resulted in a decrease in performance. Notably, the exclusion of the 'selection' step resulted in a substantial drop in performance, dipping below the outcomes achieved with the FA block. This impact was pronounced enough to cause ESIM's inference accuracy on QQP and SNLI to fall below that of their original baseline networks. Furthermore, the error bars indicate increased training instability after the removal of the "selection".

Figure~\ref{fig.ablation} illustrates the loss-epochs curves (dev.) for the ESIM model trained on two datasets. It is evident that the introduction of the FA and SFA blocks not only maintains or even improves training convergence speed but also enhances overall convergence performance. On the other hand, introducing a SFA block without the "selection" leads to numerous inefficient training processes. The loss exhibits little variation, significantly slowing down the convergence compared to previous models. As explained in Section~\ref{sec.grad}, the omission of the "selection" yields uniform gradient updates across different Inception branches during feature extraction, leading to reduced training efficiency. This obstructed backpropagation process also hinders effective convergence during training, resulting in outcomes that may even be inferior to those of the base network.


\begin{figure}[t]
  \centering
  \includegraphics[width=1\linewidth]{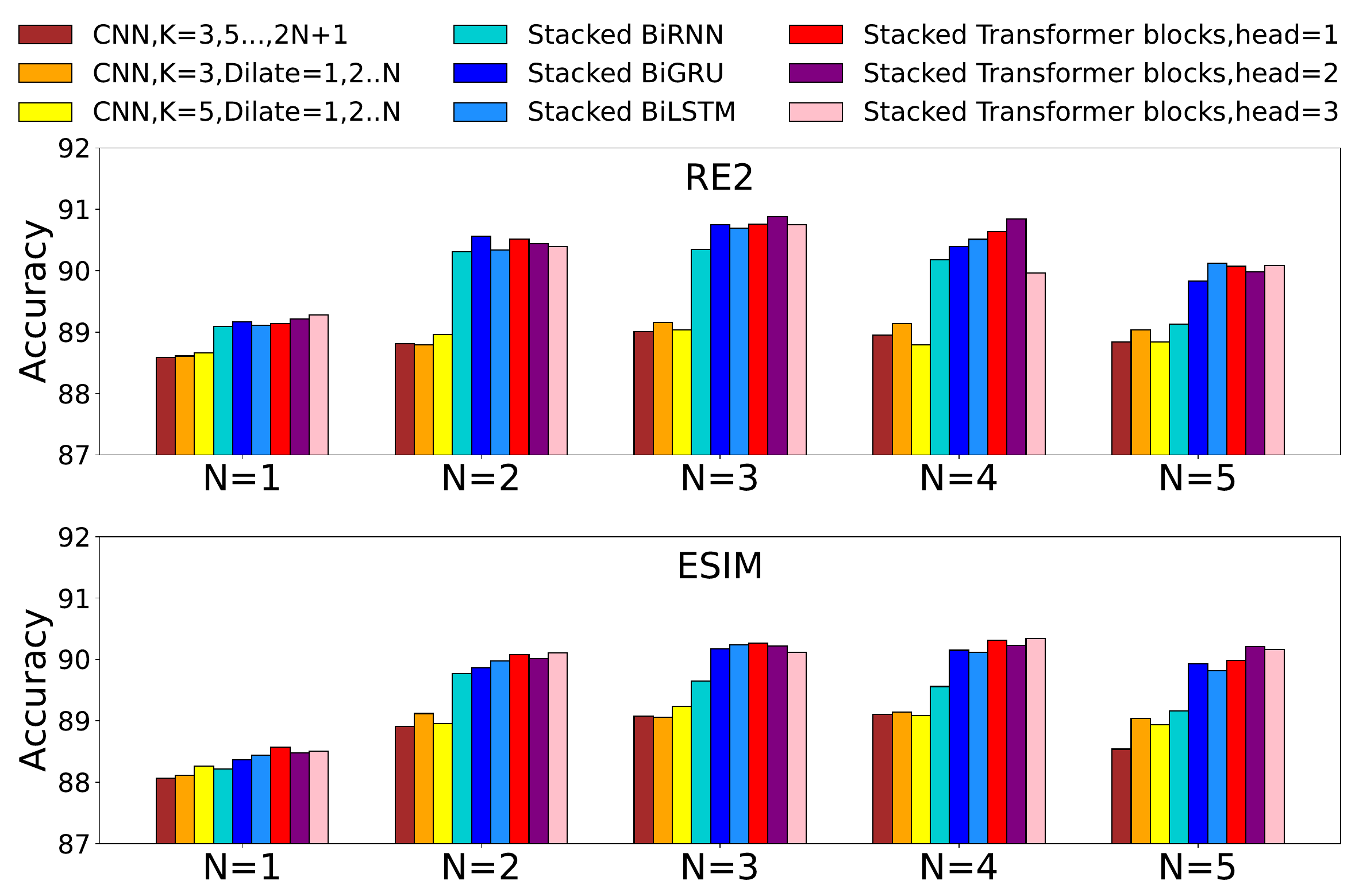}
  \includegraphics[width=1\linewidth]{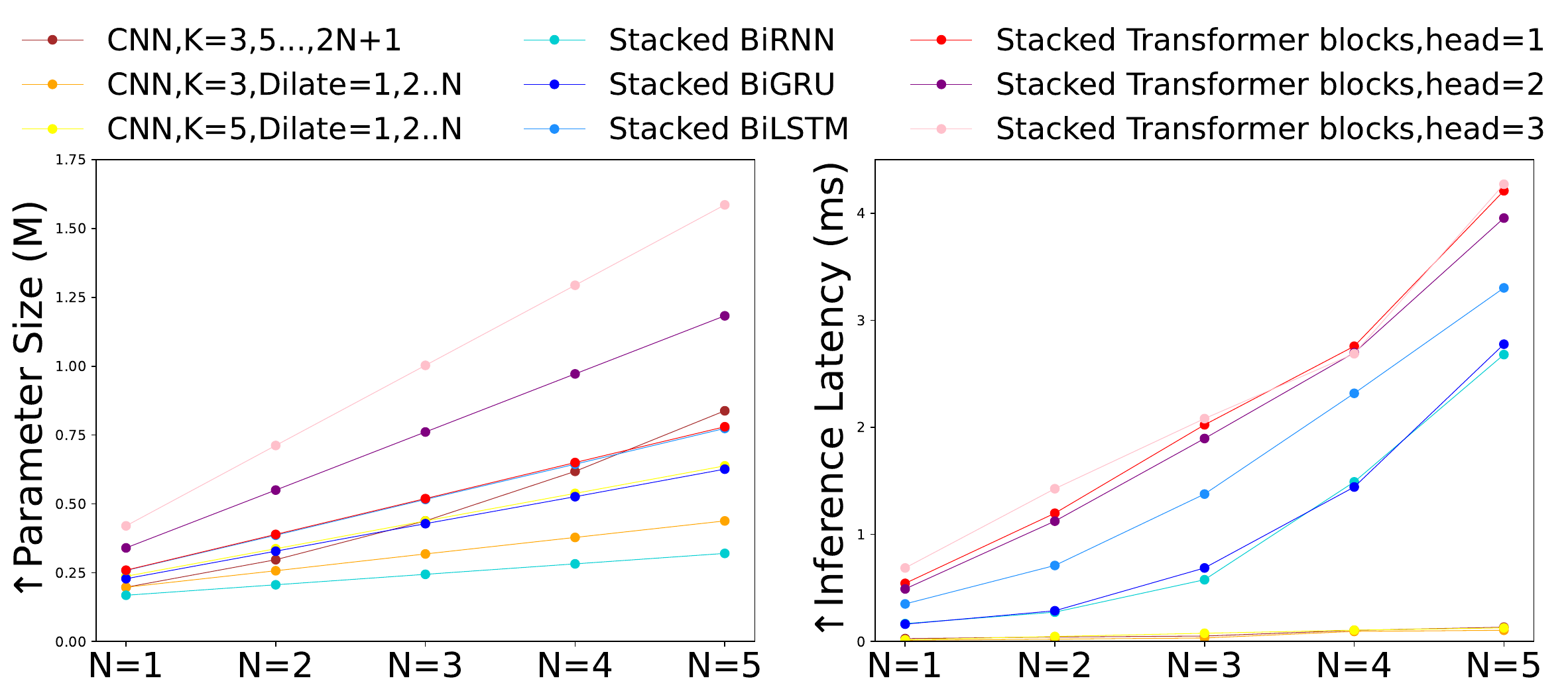}
  \caption{The average increase in evaluation accuracy (\%) of SFA blocks on QQP and SNLI with different Inception networks (using RE2 and ESIM as baselines), along with the associated parameters and average inference latency growth.}
  \label{fig.inception}
\end{figure}

\subsection{Inception Networks}

Section \ref{sec.ablation} corroborates that the "selection" is the pivotal element in the SFA block, which is built upon the foundation of multi-branch Inception networks. In this section, we discuss the effects of semantic multi-scale mapping using various Inception networks. These networks can be categorized into three types based on their fundamental architecture: CNN, RNN, and Transformer. For the CNN series, we investigate three forms: (1) CNNs with varying one-dimensional kernel sizes ($K=2N+1$) across different branches, (2) CNNs with a fixed kernel size of 3 but varying dilation factors ($\text{Dilation}=N$) across branches, and (3) CNNs with a fixed kernel size of 5 and varying dilation factors ($\text{Dilation}=N$). For the RNN series, we explore stacked BiRNN, stacked BiGRU, stacked BiLSTM. For the Transformer series, we examine three forms of stacked Transformer blocks with 1, 2, 3 attention heads, respectively. Specifically, we continue to use RE2 and ESIM as baselines and introduce SFA blocks with different Inception networks on top of them. Figure~\ref{fig.inception} illustrates the average accuracies of the networks on the QQP and SNLI datasets. A significant increase in accuracy is observed when the number of branches changes from 1 to 2 in each Inception structure, further emphasizing the criticality of the "selection" based on multi-branch Inception. Compared to the RNN and Transformer series, SFA blocks built on the CNN series do not exhibit superior performance, possibly because RNN and Transformer architectures are suited for capturing sequential characteristics. When the branch amount reaches 3, the RNN and Transformer series generally achieve the highest inference accuracy.

Figure~\ref{fig.inception} illustrates the increase in parameters and inference latency introduced by different Inception based SFA blocks, under the constraint of identical bottleneck factors ($r_1$, $r_2$). It is observed that at $N=3$, stacked BiLSTM and stacked Transformer blocks incur relatively high parameters and inference delay, contradicting the principle of lightweight text matching. In comparison, stacked BiRNN and stacked BiGRU strike a balance between performance and computational costs. SFA blocks based on stacked BiGRU exhibit superior and more stable accuracy compared to stacked BiRNN. This is the reason why we opted for a stacked BiGRU as the Inception network within the SFA block.


\subsection{Attention Analysis}


The interaction of sentence embeddings directly influences the performance of matching networks. To visually illustrate the impact mechanism of FA and SFA blocks on the network, We selected two sets of sentence pairs from the MRPC, both of which have an 'irrelevant' relationship between them: "\emph{Robin Saunders, head of the bank's London-based principal finance unit, is also expected to quit.}" \& "\emph{Robin Saunders, head of the principal finance unit, has made clear she has funding to buy parts of the business.}" and "\emph{In the second quarter, Anadarko now expects volume of 46 million BOE, down from 48 million BOE.}" \& "\emph{Production for the second quarter was cut to 46 million barrels from 48 million barrels.}".

\begin{figure}[t]
\centering
\includegraphics[width=1\linewidth]{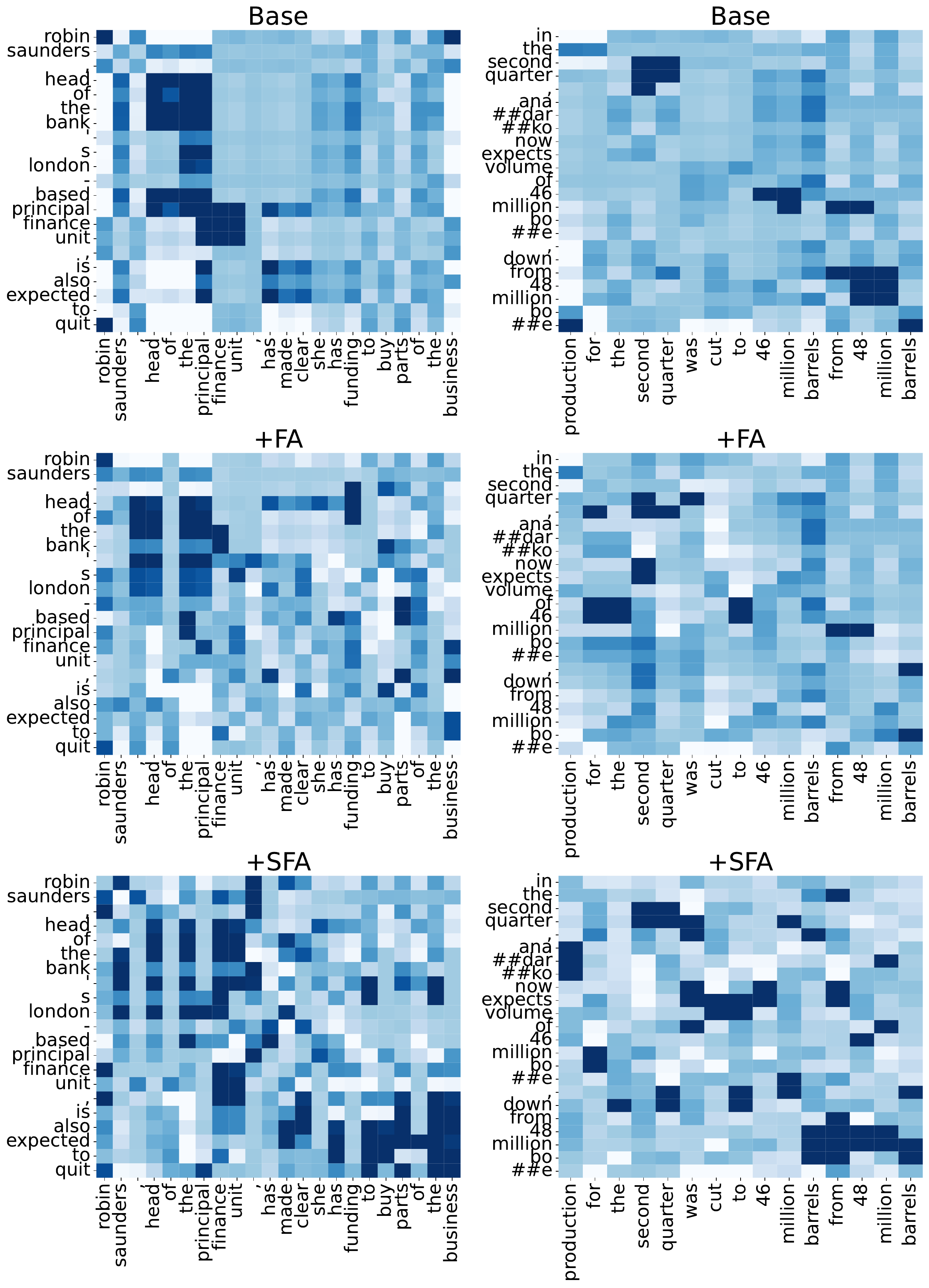}
\caption{
The heatmap of the dot product matrix for sentence pair embeddings, where deeper colors indicate higher levels of activated word-level attention.}\label{fig.case}
\end{figure}

We encoded the two sentence pairs using three types of trained ESIM (base, +FA, +SFA) respectively, and visualized the word-level dot product matrices of the base network $\bm{x}$ and $\bm{y}$, as well as those of the network $\bm{u}$ and $\bm{v}$ with FA and SFA integrated. For visualization purposes, we performed feature-level average pooling, as illustrated in Figure~\ref{fig.case}. It can be observed that the base network only activates attention between synonyms, such as "\emph{head of}," "\emph{principal finance unit}", "\emph{second quarter}", "\emph{46 million}" and "\emph{48 million}". When the FA block is introduced, this situation remains unchanged, as focusing only on synonyms does not make the network aware of their lack of relevance. In contrast, the introduction of the SFA block prompts the network to activate additional segments, such as "\emph{is also expected to quit}" \& "\emph{has funding to buy parts of the business}" and "\emph{now expects}" \& "\emph{was cut to}". These segments are essential for the network to determine the "irrelevant" relationship between sentences. This is because the FA block only adjusts the weights among individual features of each word, not affecting the values post-average pooling of all features (the weight of the sum of word-level features), and thus cannot directly influence word-level interactions. On the other hand, the SFA block aggregates embedding features across different scales with weighting. The mappings at these different scales are nonlinear, which leads to changes in the weight of the sum of word-level features. This directly activates words that were previously not focused on by the network, enabling the extraction of semantic information at a finer granularity and capturing the semantic focus.

\section{Conclusion}

In this paper, we introduce innovative attention modelling at the embedding feature level within lightweight text matching networks, presenting the FA and SFA blocks. The structures of the FA block and SFA block are concise, plug-and-play, and offer vast opportunities for expansion. In terms of structural design, beyond being a selection mechanism, the feature attention structure can support various forms, enabling finer-grained feature modeling. In terms of task format, feature attention is merely an activation of feature dependencies, unaffected by task formats. This indicates its applicability to various other semantic embedding-based tasks in NLP, such as text classification, entity recognition. We hope to encourage more researchers to explore diverse forms of feature-level attention across a broader range of NLP tasks, fostering a community ecosystem for feature attention in NLP.


\appendix

\bibliographystyle{named}
\bibliography{ijcai23}

\end{document}